\documentclass[journal,onecolumn]{IEEEtran}
\IEEEoverridecommandlockouts
\usepackage{cite}
\usepackage{amsmath,amssymb,amsfonts}
\usepackage[ruled,vlined]{algorithm2e}
\usepackage{algorithmic}
\usepackage{pgfplots}
\pgfplotsset{compat=1.14}
\usepackage{textcomp}
\usepackage{multirow}
\usepackage{booktabs}
\usepackage{adjustbox}
\usepackage{soul}
\usepackage{dirtytalk}
\usepackage{mwe,units}
\usepackage{acronym}
\usepackage{ragged2e}
\usepackage{makecell}
\usepackage{efbox,graphicx}
\efboxsetup{linecolor=black,linewidth=0.5pt}

\begin{document}

\title{A unified software/hardware scalable architecture for brain-inspired computing based on self-organizing neural models}

\author{
\IEEEauthorblockN{Artem R. Muliukov\,$^{1,*}$, Laurent Rodriguez\,$^{1}$, Benoit Miramond,$^{1}$,
Lyes Khacef\,$^{2}$, \\ Joachim Schmidt\,$^{3}$, Quentin Berthet\,$^{3}$ and Andres Upegui\,$^{3}$} \\
\IEEEauthorblockA{\textit{$^{1}$Université Côte d'Azur / Laboratoire d’Electronique, Antennes et Télécommunications / CNRS, Biot, France \\
$^{2}$Bio-Inspired Circuits and Systems Lab, Zernike Institute for Advanced Materials, University of Groningen, Netherlands \\
$^{3}$Institute of Information Technologies, Hepia, 
University of Applied Sciences and Arts of Western Switzerland, Geneva, Switzerland} \\
artem.muliukov@univ-cotedazur.fr}}
\maketitle


\begin{abstract}
The field of artificial intelligence has significantly advanced over the past decades, inspired by discoveries from the fields of biology and neuroscience. The idea of this work is inspired by the process of self-organization of cortical areas in the human brain from both afferent and lateral/internal connections.
In this work, we develop an original brain-inspired neural model associating Self-Organizing Maps (SOM) and Hebbian learning in the Reentrant SOM (ReSOM) model. The framework is applied to multimodal classification problems. Compared to existing methods based on unsupervised learning with post-labeling, the model enhances the state-of-the-art results. This work also demonstrates the distributed and scalable nature of the model through both simulation results and hardware execution on a dedicated FPGA-based platform named SCALP (Self-configurable 3D Cellular Adaptive Platform). SCALP boards can be interconnected in a modular way to support the structure of the neural model. Such a unified software and hardware approach enables the processing to be scaled and allows information from several modalities to be merged dynamically. The deployment on hardware boards provides performance results of parallel execution on several devices, with the communication between each board through dedicated serial links. 
The proposed unified architecture, composed of the ReSOM model and the SCALP hardware platform, demonstrates a significant increase in accuracy thanks to multimodal association, and a good trade-off between latency and power consumption compared to a centralized GPU implementation.
\end{abstract}

\begin{IEEEkeywords}
brain-inspired computing, neuromorphic architectures, self-organizing map, hebbian learning, multi-modal classification, post-labeled unsupervised learning, FPGA, distributed computing. 
\end{IEEEkeywords}


\section{Introduction}

\subsection{Self-organization as the main brain-inspired computational principle}

Nowadays, power consumption is one of the most crucial bottlenecks in the development of computing technologies. An important source of that inefficiency is a lack of specialization of computing systems for solving specific tasks, such as Artificial Intelligence (AI) problems. The best example of an effective system for solving equivalent problems is the human brain.
Unlike their electronic counterparts, biological neurons use direct neural connections organized in a complex three dimensional structure. The signal of interneuron axons is transmitted in an extremely specialized manner, saving on the amount of transmitted information and permitting a high scalability of the system. This behaviour appears completely different from the usual Von Neumann architecture found in a CPU. It offers universal computation capabilities but also has technical and intellectual constraints, mainly because of the physical separation between memory and computation units. 
This observation led us to develop a new computational paradigm combined with new electronic devices that support specialized neural models. 
Both are inspired by the self-organization property of the cortical areas of the biological brain, and their ability to learn their structure and function in an unsupervised manner while simultaneously acquiring data.


Self-organization can be defined as a global order emerging from local interactions \cite{heylighen2003self-organization} without a global controller or an external supervisor.
In particular, local plasticity is the fundamental computational paradigm of cortical plasticity which enables self-organization in the brain, that in its turn enables the emergence of consistent representations of the world \cite{varela1991embodied_mind}.
In addition to biological plausibility, locality can be a key insight from the brain to enable online learning in embedded (and embodied) systems. In fact, local computing implies locality in time, which satisfies the real-time constraint of online learning, and locality in space, which is a direct consequence of the co-localization of memory and computation that satisfies the energy-efficiency constraint of on-chip learning. It is therefore a major shift from standard gradient-descend learning with back-propagation paradigm. Furthermore, it is unsupervised by nature and contrasts with supervised learning using labeled data, which is impractical in most online scenarios.

Therefore, the key inspiration of this work is the functioning of the cerebral cortex and its impressive self-organizing ability \cite{cain_computational_2016}. Thanks to this ability, mammals and many higher animals create neural maps that represent their environment. Each map stores relevant information that, when combined, builds a model of the experienced objects and concepts. These maps can be built by the brain without the need for explicit data annotation or object labels. This property would be extremely useful for training on large amounts of data, where annotating itself becomes a difficult and expensive task. 
The Self-Organizing Map (SOM) \cite{kohonen1990self} is one of the main unsupervised neural models based on such principles of cortical self-organization in the brain. It dynamically adapts to represent the input data according to its distribution.
This paper explores the efficiency of the proposed brain-inspired computing software and hardware framework based on SOMs with application to the specific case of multimodal association problems.

\subsection{Multimodal association}
Most processes and phenomena in the natural environment are expressed under different physical guises, which we refer to as different modalities. Multimodality is often considered as the first principle for the development of embodied intelligence \cite{smith2005embodied_cognition}. Indeed, biological systems perceive their environment through diverse sensory channels: vision, audition, touch, smell, proprioception, etc. 
For example, we can recognize a dog by seeing its picture, hearing its bark or rubbing its fur. These features are different patterns of energy in our sensory organs (eyes, ears and skin) that are represented in specialized regions of the brain.
The fundamental concept is that there is a redundancy in neural structures \cite{edelman1987neural_darwinism}, known as degeneracy, which is defined as the ability of structurally different biological elements to perform the same function or yield the same output \cite{edelman2001degeneracy}. In other words, it means that any single function in the brain can be carried out by more than one configuration of neural signals, so that the system still functions after the loss of one component. It also means that sensory systems can educate each other, without an external teacher \cite{smith2005embodied_cognition}.

The same principle can be applied for artificial systems, since information about the same phenomenon in the environment can be acquired from various types of sensors: cameras, microphones, accelerometers, etc. Due to the rich characteristics of natural phenomena, it is rare that a single modality provides a complete representation of the phenomenon of interest \cite{lahat2015multimodal_overview}. Therefore, an important problem in the development of modern AI systems consists in the aggregation of data from different sources to obtain a more complete and homogeneous understanding of the world, known as multimodal learning. In the brain, the representations built by different zones of the cortex might complement each other, for example by direct signal exchange \cite{cappe2009multisensory}. This complementarity between sensory maps develops and relies on the self-organization mechanism of lateral synaptic communication. To simulate this mechanism, we use in this paper Hebb's learning principle: "neurons that fire together, wire together" \cite{hebb-organization-of-behavior-1949}. It suggests reinforcing the connections between co-activating neurons to capture the co-occurrence of specific data representations among different modalities caused by the same phenomenon.

Recent works have tried to study the human brain’s ability to integrate inputs from multiple modalities \cite{calvert2001crossmodal_processing, kriegstein2006multisensory_association}, and it is not clear how the different cortical areas connect and communicate with each other. To answer this question, Edelman proposed the reentry framework \cite{edelman1982reentry, edelman1993neural_darwinism}. Reentry is a process which involves a population of excitatory neurons that simultaneously stimulates and is stimulated by another population \cite{edelman2013reentry}. For example, it has been shown that reentrant neuronal circuits self-organize early during the embryonic development of vertebrate brains \cite{singer1990assemblies, shatz1992thalamus_cortex}, and can give rise to patterns of activity with Winner-Take-All (WTA) properties \cite{douglas2004neocortex}. When combined with appropriate mechanisms for synaptic plasticity, the mutual exchange of signals amongst neural networks in distributed cortical areas results in the spatio-temporal integration of patterns of neural network activity. It allows the brain to categorize sensory inputs, remember and manipulate mental constructs, and generate motor commands \cite{edelman2013reentry}. Thus, reentry may be the key to multimodal integration in the brain.
Based on the reentry paradigm, we have previously proposed the Reentrant SOM (ReSOM) model \cite{Khacef2020_ReSOM}, a self-organizing artificial neural network based on local plasticity mechanisms for unsupervised learning and multimodal association \cite{Khacef2020_PhDthesis}.
In this work, we significantly upgrade it by enhancing its scalability and providing a first hardware implementation.

\subsection{Contributions of the paper}
This work adopts the use of direct connections between SOMs to simulate the process of self-organization among areas of the cerebral cortex. This mechanism is proposed to solve basic AI problems such as clustering, classification or anomaly detection. A distinctive feature of our algorithm is its scaling capability and the ability to train neural connections in the absence of direct data annotation. The architecture was tested in the presence of numerous data modalities, and the results of our model surpass the accuracy of previously published models  \cite{Khacef2020_ReSOM, rathi_stdp_2021} by several percent, bringing us extremely close to real-life applications.

The model was implemented in the form of scalable constructors combining several digital devices (SCALP boards introduced in section \ref{sec:SCALP}) connected through high-speed serial links (HSSL). Based on the results of our previous work \cite{Khacef2020_ReSOM}, this paper proposes an extension of the existing framework that enriches it with new tests and offers an implementation both in software for simulation and in hardware for real-time prototyping. This contribution presents a unified software/hardware architecture inspired by cortical self-organization and thus makes an important step towards embodied "brain-like" electronic systems.
Such a model could be used in the future by an automated embedded system (robot, drone, autonomous car, space rover, etc.) to build a representation of its surrounding environment through multimodal sensors.

\subsection{Outline of the paper}
The next section presents an overview of the state of the art related to the scientific questions addressed in this paper. Section \ref{sec:ReSOM} describes the ReSOM model for multimodal unsupervised learning based on local computations. Section \ref{sec:SCALP} gives general information about the SCALP board used for the hardware solution. Section \ref{sec:Results} describes both software simulations on a CPU/GPU and the results of hardware deployment on FPGA-based boards. Section \ref{sec:Discussion} briefly summarizes the tests conducted and discusses possible paths for future development.


\section{State of the art}
\label{sec:SotA}

\subsection{Brain-inspired self-organizing neural models}
As observed in the brain, the product of the {\bf self-organization} process is a representation of the stimulus that is feeding this structural self-organization, joining together the structure and the function. This ability to extract information from experience and interaction both with the environment and with others is observed especially in living systems. In our work we pay particular attention to analogous self-organizing models with the goal of reproducing the brain's unique learning qualities. Regardless of the nature of the  self-organizing model being considered, it must satisfy several properties that we seek to achieve for embodied brain-inspired computing: 
\begin{itemize}
    \item Capability of distributed computing (for hardware implementation and scalability purposes);
    \item Capability of unsupervised learning;
    \item Capability of multimodal data processing.
\end{itemize}

These specific behavioural needs point the way to another form of unsupervised learning compared to the approaches proposed in classical Machine Learning (ML). In this review, we do not go too deeply into the details of either the Auto-Encoders (AE) or the Semi-Supervised Learning (SSL) or  Multi-Agent (MA) systems, as they involve relatively similar but at the same time quite distinct problems and restrictions compared to ours. We also do not try to cover all the existing self-organizing and unsupervised methods, considering the enormous number of publications in this domain. So, we mention here only some of the algorithms that have direct overlap with the unique features of our framework.

A distinctive feature of our model is the construction of an explicit data representation map of the objects. It is worth mentioning that other solutions tackling this issue have been proposed in the literature, although none of them meet all our criteria. 
One of them is the Elastic Map \cite{gorban2005elastic} method that creates a nonlinear distortion of the observed space, and another is the Generative Topographic Map \cite{bishop1998gtm} that builds a map based on statistical principles, without an inspiration from neural systems. Separately, we note the Neural Gas (NG) model \cite{martinetz1991neural}, which works on similar principles as the SOM, but allows more freedom in the neural structures achievable. The NG model on the other hand, lacks the topological properties inherent in cortical regions, since its structure depends on the distance of each prototype from stimuli. This makes it complex to implement based only on local interactions of neurons, therefore complex to program in a distributed manner.

Also important in the context of self-organizing systems are Spiking Neural Networks (SNNs) based on local synaptic plasticity, such as Spike-Timing Dependant Plasticity (STDP) \cite{bichler2012pcm-synapse,diehl_unsupervised_2015, Hazan_IJCNN_2018, rathi_stdp_2021}. These models exploit the temporal dynamics and the event-based computing of spiking neurons. 
Both SOMs and SNNs show interesting properties like {\bf distributed computing} and {\bf unsupervised learning}, but at the cost of a massive number of interconnections between neurons. In fact, even though the learning process of the SOM \cite{kohonen1982som} and SNN \cite{diehl_unsupervised_2015} can be distributed with local computing, the competition mechanism for the emergence of the "winning neuron" is usually implemented using either a centralized unit or an all-to-all connectivity between neurons. These two approaches are not scalable in execution time and connectivity, respectively, as further discussed in \cite{Khacef2019_SOMs}. 
In order to overcome this limit, we proposed in \cite{rodriguez2018ig_som} to distribute the SOM computing based on the Iterative Grid (IG), a cellular neuromorphic architecture with local connectivity amongst neighbor neurons. This novel implementation gives a much better scalability of the SOM in terms of neurons. The IG is not used in this work, though, because our objective is to demonstrate the scalability of ReSOM in terms of SOMs, where every map represents a given modality.

This work is therefore based on the Kohonen's SOM (KSOM), which is at a good level of biological abstraction for our application and has a number of advantages over the previously mentioned models. Typically, its advantages include all the previously cited desired properties, including ease of hardware implementation. Indeed, the presence of a strict network structure endowed with a local neighborhood simplifies its further implementation in electronic circuits such as FPGA or ASIC.
In parallel, it has been shown that SOMs perform better at representing overlapping structures compared to classical clustering techniques such as partitive clustering or K-means \cite{budayan2009cluster_vs_som}. Recently, we showed that the SOM combined with transfer learning reaches a competitive accuracy on complex few-shot learning problems \cite{Khacef2020_FewShotGPU}.
In addition, SOMs were directly inspired by work on the cerebral cortex \cite{kohonen1990self}, and can hence be potentially applied in the future for the development of biologically compatible components.

The KSOM proposed more than thirty years ago \cite{kohonen1991self} can serve as the main reference for our work, even though different adaptations of the original SOM model have since been proposed. We can distinguish several models offering some interesting and unique behaviors. The Dynamic Self-organizing Map (DSOM) \cite{rougier2011dynamic}, for example, dynamically adapts to changes in the dataset over time, continuously rebuilding the map structure if needed. The Growing SOM (GSOM) \cite{dittenbach2000growing} and the Plastic SOM (PSOM) \cite{lang2002plastic} can change the network structure by adding/deleting neurons during the learning process. So it becomes possible to dynamically adapt the size of the network to the data structure, but at the cost of a more expensive learning process. The C(ellular) SOM model (CSOM) \cite{girau2019cellular} also offers gains in accuracy and energy consumption, using a simplified grid to enhance its hardware implementation. The Pruning Cellular Self-Organizing Maps (PCSOM) model \cite{upegui2018pruning} prunes some SOM connections to gain better performance. Furthermore, certain models such as the Semi-Supervised SOM propose to aggregate mixed supervised and unsupervised data to achieve better representations with the help of known labels \cite{braga2020deep}.

Despite these possible extensions of the model, this work refers to the original KSOM version. This model is sufficient to prove the concepts addressed in this paper. Moreover, we have shown that the standard KSOM method gives a better accuracy in classification tasks using post-labeled unsupervised learning \cite{Khacef2019_SOMs}. Nevertheless, this choice does not exclude the possibility of using the unified framework proposed here with other versions of self-organizing cortex-inspired maps in future studies, depending on the problems being addressed.

\subsection{Brain-inspired multimodal frameworks}
Multimodal interaction is another key aspect of this paper. Taking into account the fact that each SOM is associated with the data of one modality, it can be assumed that each neural map acts as a region of the cerebral cortex. 
Our model uses Hebbian connections \cite{hebb-organization-of-behavior-1949} to transfer activity signals between two of these maps. 
By behaving in this way, the cortex is able to correct the weakness of one modality region using another. The proposed framework tries to achieve the same behaviour, simulating the interaction among cortical zones of the human brain. The relevance of such a method is confirmed by studies in cognition that show the similarity of this training method to the developmental learning observed in children in the first years of life \cite{althaus2013modeling}.
But to our knowledge, no computational model of this cognitive scheme has been proposed before. And moreover, no electronic version has been designed.
The only work tackling this goal is ReSOM, a model proposed in \cite{Khacef2020_ReSOM} on which this paper is largely based. Specifically, this work is an extension of the ReSOM, keeping its general principles, but adding new features for scalability and a hardware implementation, as explained in the following.

A considerable number of previous works have been devoted to analyses of the interactions between self-organizing maps for the purpose of studying their communication \cite{lallee_multi-modal_2013, 10.3389/frobt.2016.00022, lefort2010self, morse2010epigenetic}. Furthermore, most such studies have already tested the aggregation of several modalities, but with important differences from this work. In particular, the models were used to solve other problems, for example the ones found in developmental robotics (orientation of robots in space, sensory-motor coordination, etc.), and not the clustering problem targeted in this paper. Also, the connection configurations were quite different and included an extra region for the modalities, resembling the so called Convergence-Divergence Zone (CDZ) \cite{lallee_multi-modal_2013}. A distinctive feature of the CDZ-based method is the use of an additional map to connect all the SOMs together to combine their activities. The present work follows another paradigm and explores the retracing of direct reentry links between several neural maps, making it applicable to another type of AI problem (such as the clustering problem).
Some authors propose a solution to the clustering problem  \cite{jayaratne2021unsupervised}, but with a focus on the field of big data computing with map-reduction (map-reduce) approach. It uses the Apache Spark \cite{10.1145/2934664} framework and moves away from a biological inspiration.

Several other works in the literature have proposed to combine multimodal data without the direct use of self-organizing mechanisms. These include the following: \cite{cholet2019bidirectional} proposed to use the Bidirectional Associative Memory; \cite{parisi2017emergence} proposed hierarchical architecture with Growing When Required (GWR) networks, where action–word mappings are developed by binding co-occurring audiovisual inputs using bidirectional inter-layer connectivity; \cite{nakamura2011bag} proposed to use a joint probability with the Latent Dirichlet allocation model; and \cite{vasco2020mhvae} proposed to enforce the classification by taking the Variational Encodings as a part of their framework. All these methods offer different points of view on the question of the representation of information in AI. They provide neither a midterm global framework which cognition tasks could rely on, nor an obvious short-term advantage over the model studied here. But we will monitor their evolution as possible competing models to our framework. 

There is also a huge research area investigating multimodal encoders based on conventional neural networks. They achieve state-of-the-art results for solving particular machine learning problems. It is practically impossible to analyze all of them, but a significant number of them, at one step or another, use the classical concatenation of multimodal vectors \cite{bendre_generalized_2021, chen_multimodal_2020, xie_multimodal_2020}, without a deep examination of unique dependencies between them. Nevertheless, there are other models proposing smarter modality aggregation, such as the Contrastive Multimodal Fusion method \cite{liu2021contrastive}, showing there is growing interest in the ML community for nontrivial multimodal fusion. However, they are still far from implementation in electronic circuits, especially for computation on separate and scalable electronic devices, as is proposed in this work.

Each method mentioned in this section may be placed in an overall compromise between performance and energy efficiency. It should be remembered that the energy cost of machine learning methods comes much more from the learning phase than from the inference phase. \cite{brownlee2021exploring}
This aspect is often overlooked in even the most recent neural accelerators.
Taking this issue into account from both a software and a hardware point of view is therefore essential for the more general deployment of AI in the future. 

\subsection{Neuromorphic (electronic) architectures}

Neural networks implementations in the the form of hardware architectures haves gained an increased interest in recent years. The recent works around Deep neural networks for pattern recognition have put a spotlight on neuromorphic engineering and some proposed architectures, like  \cite{Paindavoine:2015} or \cite{Lecun2012} have demonstrated the computation efficiency of neuromorphic systems in terms of energy consumption. More generally, neuromorphic engineering aims to emulate as many neurons and synapses as possible with dedicated architectures in digital circuits as \cite{DANNA2015} or analog systems as \cite{indiveri2011frontiers}.

Leading  projects in neuromorphic engineering have led to the creation of powerful brain-inspired chips capable of emulating multiple spiking neurons to investigate a new type of computing architecture to aid neuroscientists' research or pave the way for efficient embedded AI. For example, as part of the DARPA's SyNAPSE project roadmap, the IBM's TrueNorth neuromorphic chip (\cite{Merolla668})(\cite{cassidy2013cognitive}) can implement 1 million digital neurons. The ThrueNorth chip implements a very rich neural model, capable of reproducing many behaviors observed in biological neurons. 
As part of the Eureopean Human Brain Project, the SpiNNAker (\cite{jin2010modeling}) project aims to model a billion biological real-time impulse neurons using a million ARM968 cores. More recently, Intel's Loihi neuromorphic chip (\cite{davies2018loihi}) implements 130K Leaky Integrate and Fire (LIF) neurons and 130M synapses capable of online learning and inference. Its successor Loihi2 (\cite{ref_loihi2}) reaches 1M neurons and introduces a more flexible microcode programmable neural engine that allows simulation of a wide range of different neural models.

Though technologically impressive, these chips are designed for neural network simulation, not for the self-organization of hardware resources and are not intended to be used for SOMs models.

Self-organizing neural networks have also received their share of interest and several works have been published on the subject of dedicated SOM hardware architectures. As SOMs can be computational expensive if implemented in a straightforward manner a lot of efforts have been invested to propose efficient dedicated hardware SOM architectures. Most of the existing works focus on optimizing the time and power performances by modifying the KSOM algorithm to new "hardware-friendly" variants, less expensive in term of hardware resources such as logic and memory elements. Simplification of numerical representation (fixed point instead of float point) and/or redefinition of functions (Manhattan distance instead of Euclidean) are common optimization techniques aiming to optimize performance incurring in the cost of loosing algorithm accuracy (\cite{4099981, Younis2009ReconfigurableSN, 6909349, 8050799, 7966351}). More original vector codings have been proposed by \cite{HIKAWA2005514} which proposes to encode the vector components as the duty cycle of square waveforms and to use Digital Phase-Locked Loops (DPLL) as parallel computing elements to measure and update the distances between vectors.

Other approaches focus on learning performance (expressed in millons of connection updates per second - MCUPS - per watt consumption). Lachmair et al. (\cite{Lachmair_2013}) propose a bus-interconnected multi-FPGA hardware SOM that allows the implementation of large reconfigurable networks. Abady et al. propose a layered description of the HW SOM (\cite{Abadi_2016}, \cite{Abadi_2018}) where the computational neurons are decoupled from the communication that is implemented with a NoC. The same idea is exploited by Jovanovic et al. (\cite{Jovanovic_2018}, \cite{iet:/content/books/10.1049/pbpc034e_ch16}) in a more hierarchical manner as the interconnected computational cells no longer implement neurons but clusters of neurons. Each cluster computes a local winner and the global winner is computed through the NoC in a systolic way. 


This interest is well-intended as efficient hardware architectures for self-organizing models will enable their application to autonomous and embedded systems, where unsupervised on-line learning is of first importance. Nonetheless, there is no dedicated SOM architecture in the literature offering the flexibility required for multimodal SOM acceleration. 


\section{Reentrant SOM, a brain-inspired model for multimodal unsupervised learning}
\label{sec:ReSOM}

In this section, we describe the neural model, which enables the fusion of multiple modalities in a scalable and unsupervised manner.
The model is an extension of the previously published ReSOM model \cite{Khacef2020_ReSOM} that only considered two modalities. 
Therefore, we start this section by defining the principles behind the KSOM, next we briefly discuss the training of intermodal (Hebbian) connections, and finally we put all these elements together to develop our neural framework for reentry multimodal interaction. This framework acts as an unsupervised method of clustering high-dimensional data that can then be used to solve several kinds of machine learning tasks. We finish this section with an application of the advanced ReSOM model to the specific case of classification tasks with numerous data sources.

\subsection{Kohonen's Self-organizing Map}
\begin{figure}[ht]
    \centering
    \includegraphics[width=0.7\linewidth]{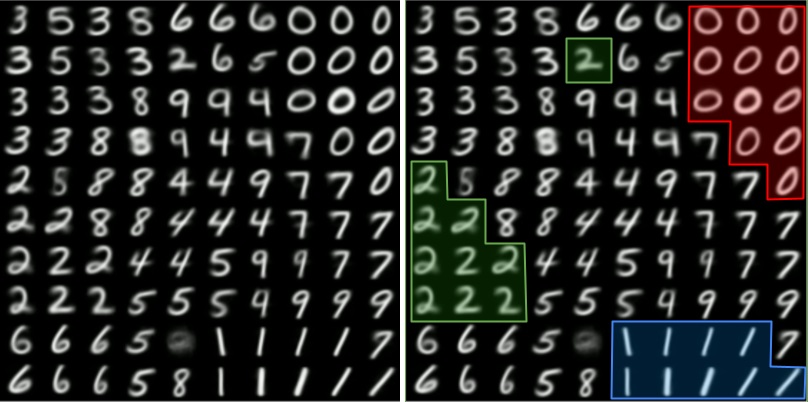}
    \caption{Example of self-organization of SOMs on the MNIST dataset. The SOM neuron weights are drawn on the image (reshaped to the form of image, to be visually clearer). On the left - only original weights, on the right - with some clusters selected.}
    \label{fig:mnist_som_repr}
\end{figure}

The model used here \cite{kohonen1991self} is a neural network consisting of a rectangular 2D grid of neurons. Each neuron is endowed with a weight vector with the same dimension as the processed data. The neural network training process consists of the following stages:
\begin{itemize}
    \item 
0. Read a vector of multidimensional data
    \item 
1. Identify the closest neuron to the current input in the grid by computing and comparing L2 distance. In the following, this neuron will be called Best Matching Unit (BMU)
    \item 
2. Train the network by changing the weights of the BMU and its neighbors in the direction of the data vector by the following formula

\begin{equation}
w_n(t+1) = w_n(t) + \epsilon(t) \times h_\sigma (t,n,s) \times (v - w_n(t))
\end{equation}
where:
\begin{equation}
h_\sigma (t,n,s) = e^{-\frac{||p_n-p_s||^2}{2\sigma(t)^2}};
\end{equation}

\begin{equation}
\sigma(t) = \sigma_i(\frac{\sigma_f}{\sigma_i})^{\frac{t}{t_f}};
\end{equation}

\begin{equation}
\epsilon(t) = \epsilon_i(\frac{\epsilon_f}{\epsilon_i})^{\frac{t}{t_f}};
\end{equation}

With \(s\) - the BMU index in the grid, \(v\) - the data vector, and \(n\) - the index of any neuron. \(\epsilon(t)\) and \(\sigma(t)\) are decreasing functions to ensure the annealing of the learning; \(p_k\) - coordinate  for a neuron with index k; \(h_\sigma (t,n,s)\) - the neighborhood function, \(t\) - "time" or learning iteration number. \(i\) and \(f\) refer to the initial and final values.
    \item 
3. Repeat the operations (0, 1 and  2) for each data sample

\end{itemize}
As a result, the grid of neurons forms a 2D representation of a real data space. Here, each neuron represents a group (or a class) of the previously shown objects, by being itself a typical (more precisely - averaged) data vector. 
Further division of the space of the representation neurons into known classes occurs, if necessary, when analyzing the associated labels. So, for example, the number 4 can be clearly represented as a closed and open figure (both in printing and writing styles). The drawing of clear boundaries between clusters is carried out in the next steps of the algorithm. But, for example in the simple MNIST case, as shown in Figure \ref{fig:mnist_som_repr}, the regions corresponding to the clusters attached to each number ("0", "1", etc.) can already be clearly observed.

\subsection{Hebbian connections}
\begin{figure}[ht]
    \centering
    \includegraphics[width=0.7\linewidth]{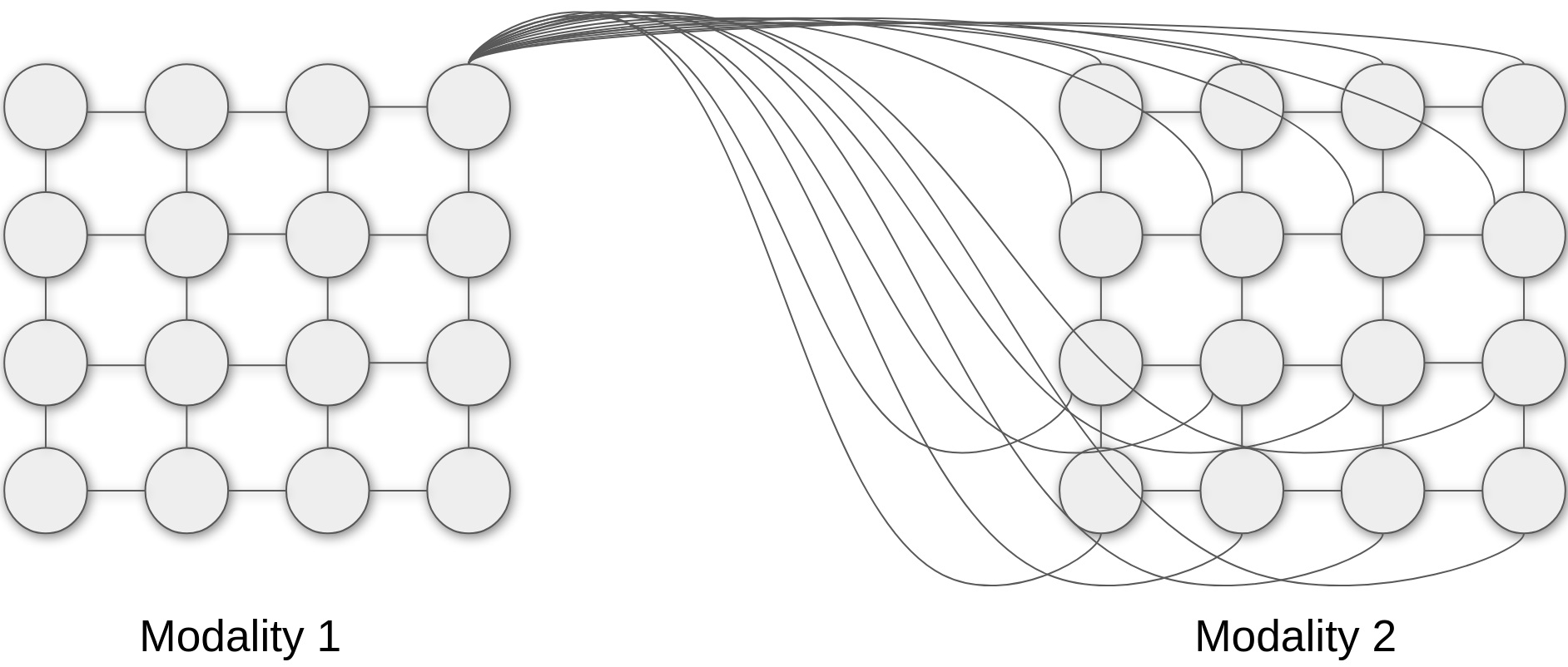}
    \caption{Schematic Hebbian connections for one neuron of a sensory map connected to another sensory map.}
    \label{fig:som_connect}
\end{figure}

To model the interaction between the zones of the cortex, it is proposed to use the model of pairwise connections between SOM maps. Each pair of neurons belonging to different SOMs is connected with a specific lateral synapse (as in Figure \ref{fig:som_connect}). During simulations, the weights of the connections between two maps are stored in a full-size matrix, with dimensions \(d_1^2\times d_2^2\), where \(d_1\) and \(d_2\) are the widths of each map.

The Hebbian learning is conducted after that of individual SOMs. The Hebb principle is used to train the lateral weights. So the neural connection between two BMUs, with indices (\(x\) and \(y\)), belonging to different neural maps is strengthened according to the following law

\begin{equation}
{W'}_{BMU_x,BMU_y} =  {W'}_{BMU_x,BMU_y} + \mu \times a^x \times a^y,
\end{equation}

where \(\mu\) is a hyperparameter and  \(a^s\) is the maximal activation function coming from a SOM with index  \(s\) (activation of the BMU):

\begin{equation}
a^s = \max_{n=0}^{k-1}{e^{-\frac{||v^s - w^s_n||}{\alpha}}}
\end{equation}

Activations are counted and used for all existing vectors \(v_s\), where \(s\) is the modality index. Thus, this operation must be repeated for all pairs of data samples representing different modalities.

\subsection{ReSOM structure and learning}

In order to scale the framework and be able to process more than 2 modalities, it is proposed in this work to:

\begin{figure}[ht]
    \centering
    \includegraphics[width=0.3\linewidth]{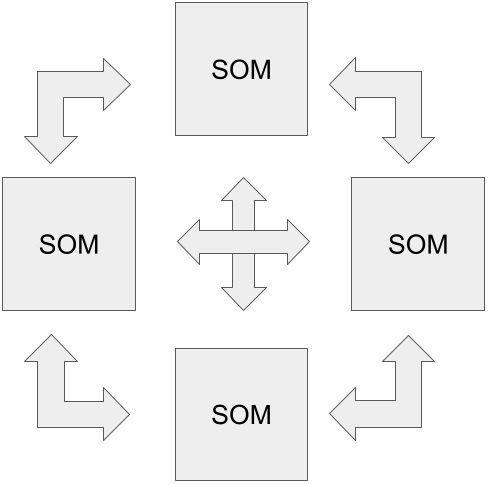}
    \caption{The ReSOM framework scheme with 4 modalities. Each double arrow corresponds to the Hebb weights between SOMs, and all are trained independently of each other.}
    \label{fig:mmresom_chema_cut}
\end{figure}

\begin{itemize}

\item
1.1. Create the number of SOMs required to process all modalities (one map per modality)
\item
1.2. Create the matrix of Hebbian weights for all pairwise compositions of all available SOMs (create \(\frac{k(k-1)}{2}\) Hebbiab connections for \(k\) maps). In Figure \ref{fig:mmresom_chema_cut} an example of interaction between 4 modalities is depicted.
\item
2. Train each SOM separately.
\item
3. Using the received SOM activations, train all the \(\frac{k(k-1)}{2}\) intermodal connections independently.
\end{itemize}

Thus, after training, we get a network of matrices representing the SOMs themselves and the connections between them, as in Figure \ref{fig:mmresom_chema_cut}.

\subsection{Application of ReSOM to classification tasks}
In order to make a quantitative analysis of our proposed model, we must use a quality metric of the resulting space clustering. To do that, we apply our framework to three subsets of data: a training set used for unsupervised learning composed of multimodal unlabeled data; a small annotated dataset used to attribute labels to neurons; and a testing set used for the validation and the evaluation of the clustering. This specific validation process leads at the end to the evaluation of a single metric corresponding to the accuracy of classification. That also means that this application consists in mapping the continuous space representation onto the discrete label space by assigning some class labels to the representation neurons. To do this, we use a certain amount of labeled data. 

In fact, for any classification task based on unsupervised learning where the results need to be communicated in the so-called discrete label space, we need to find a method of attributing the identified clusters to existing labels. Different methods may be used to solve this problem, such as having an external expert who manually labels the obtained representations, or an automated process based on some labeled data. We use the latter approach, trying to minimize the number of required labels. The labeled data space should be large enough to represent all possible patterns of the available data. In practice, our results showed that only 1\% of data for the MNIST dataset and 10\% for the SMNIST dataset is enough to reach the best quality of classification \cite{Khacef2020_ReSOM}. To have a uniform structure this work uses about 10\% of labels for all the data modalities (precise numbers are given in section \ref{sec:Results}). In the following sections, we discuss further the processes of labeling the neurons and testing and assessing the quality of a trained model.

Lastly, it should be noted that the problem addressed by our framework does not correspond to a typical SSL problem, since the existing labels are available (or provided) only after the training. So we consider a special type of SSL problem, called in the literature the Post-Labeled Unsupervised Learning problem \cite{Khacef2020_FeatureSOM, Khacef2020_FewShotGPU}.

\subsubsection{Multimodal labeling}
The first step is the labeling process. In this work the objective is to deal with numerous modalities of data. The method is based on calculating the probability for each neuron to be assigned to a class. This probability is calculated as the sum of the neuron activations over all available annotated data for each separate label (as in Equation \ref{eqn:proba_of_neuron} below, for a SOM with index \(j\) and a neuron with index \(s\)). For each neuron, its activation consists of its afferent activity added to the activations coming from the lateral connections (Equation \ref{eqn:labeling_activation}, where \(W\) are the weights of the lateral connections).

\begin{equation}
\label{eqn:proba_of_neuron}
p(j,s,label=i) =\frac{\sum_{label=i}{A_{s}}}{\sum_{label=any}{A_{s}}}
\end{equation}

\begin{equation}
\label{eqn:labeling_activation}
{A_{s}^j} = {c_1^{j}a_{s}^j} + \sum_{for l \neq j}{c_2^{l,j}a^l \cdot W^{l,j}_s}
\end{equation}

The coefficients \(c_1^{l,j}\) and \(c_2^{l,j}\) are hyperparameters, which may have any non-negative values and define the relative importance of each modality during the labeling process. These parameters define the chosen labeling mode:

\begin{itemize}
    \item If \(c_2^{l,j} = 0\) for {\bf all} \(l\) and \(j\), the mode is {\bf unimodal labeling}. Here we label only using the individual SOM activation. This method was generally used in this work, as it is the simplest one;
    \item If \(c_2^{l,j} = 0\) for {\bf some} \(l\) and \(j\), the mode is {\bf mixed labeling}. Here we label the neurons using their individual activations and some of the lateral activations;
    \item If  \(c_1^{j} = 0\) for {\bf all} \(j\), the mode is {\bf divergence labeling}. Here we label the neurons of one SOM using the activations of the other modalities. For example we can label the audio modality using the data of the visual one. Previous studies have demonstrated that using this method we can reduce both the required quantity of annotated data and the number of annotated modalities without a loss in performance \cite{Khacef2020_ReSOM}. In this paper we do not explore this possibility, since optimizing the labeling is not the focus of this study.
\end{itemize}.

\subsubsection{ReSOM testing and accuracy evaluation}
Next, we discuss the method for testing the trained model. In the process of class prediction, we count the activations a little differently than we do in the neuron labeling process.  Here, the final ReSOM activation of each neuron consists of the product of its afferent activation and all the activations coming from the lateral SOMs, according to Equation \ref{eqn:testing_activation}. 

\begin{equation}
\label{eqn:testing_activation}
{A_{s}^j} = {a_{s}^j} \times \prod_{l \neq j}{a^l \cdot W^{l,j}_s}
\end{equation}

\begin{equation}
\label{eqn:find_max_resom}
j_{max},s_{max}= arg\max_{j,s}{A_{s}^j}
\end{equation}

Next, we find the neurons, \(s_{max}\), with the maximum activations among all the  SOMs with index \(j\) using Equation \ref{eqn:find_max_resom}. We propose to calculate the maximum activation among all available neurons of all available SOMs to choose the winning neuron (which gives the final prediction). 

To evaluate the model we need to predict labels for all the test samples. The class of the selected neuron is compared with the ground truth. The accuracy is computed as the proportion of all the correctly predicted labels over the total number of test samples.


\section{SCALP: Self-configurable 3D Cellular Adaptive Platform }
\label{sec:SCALP}
\begin{figure}[ht]
    \centering
  \includegraphics[width=0.7\linewidth]{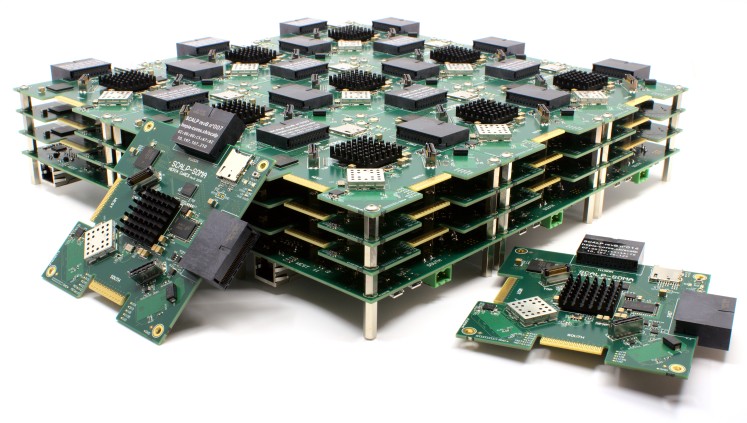}
  \caption{Array of 27 interconnected SCALP nodes.}
  \label{fig:scalp_array}
\end{figure}

Network-on-chip is a natural evolution of the increasing complexity of system-on-chip architectures. In order to cope with prototyping requirements of such complex systems we have built a 3D multi-FPGA platform called SCALP~\cite{SCALP}.
SCALP is intended to provide flexible reconfigurability and 3D interconnectivity of its basic computation nodes. Each node is a PCB (printed circuit board) mainly containing an FPGA and HSSL-based connections that allow it to connect to its six neighbors (north, south, west, east, top, bottom). Figure~\ref{fig:scalp_array} shows an array of 3x3x3 interconnected SCALP nodes.

Each SCALP node has a size of 10x10~cm and is implemented in a 12-layer PCB. Programs are executed by a Xilinx Zynq SoC (with a dual-core ARM Cortex-A9 processor @866~MHz and Artix-7 programmable logic with 74,000~cells), with its associated memory and enhanced communication capabilities permitting connections to neighbor modules through the HSSL with data rates up to 6.25~Gb/s.

A layered hardware/software architecture enables applications to be deployed on SCALP in a seamless manner. The ARM Cortex-A9 runs a Petalinux OS allowing users to write applications in C, C++ or Python. A set of libraries and drivers enables the use of services deployed on the FPGA. These services may be hardware accelerators or specific interfaces. In the case of the work presented in this paper, they will permit access to the router built on the FPGA. Figure~\ref{fig:scalp_router} depicts the SCALP node internal architecture in the case of a 2D SCALP array (a third dimension is also supported). It is composed of two main layers, routing and computation:

\begin{figure}[ht]
    \centering
  \includegraphics[width=0.7\linewidth]{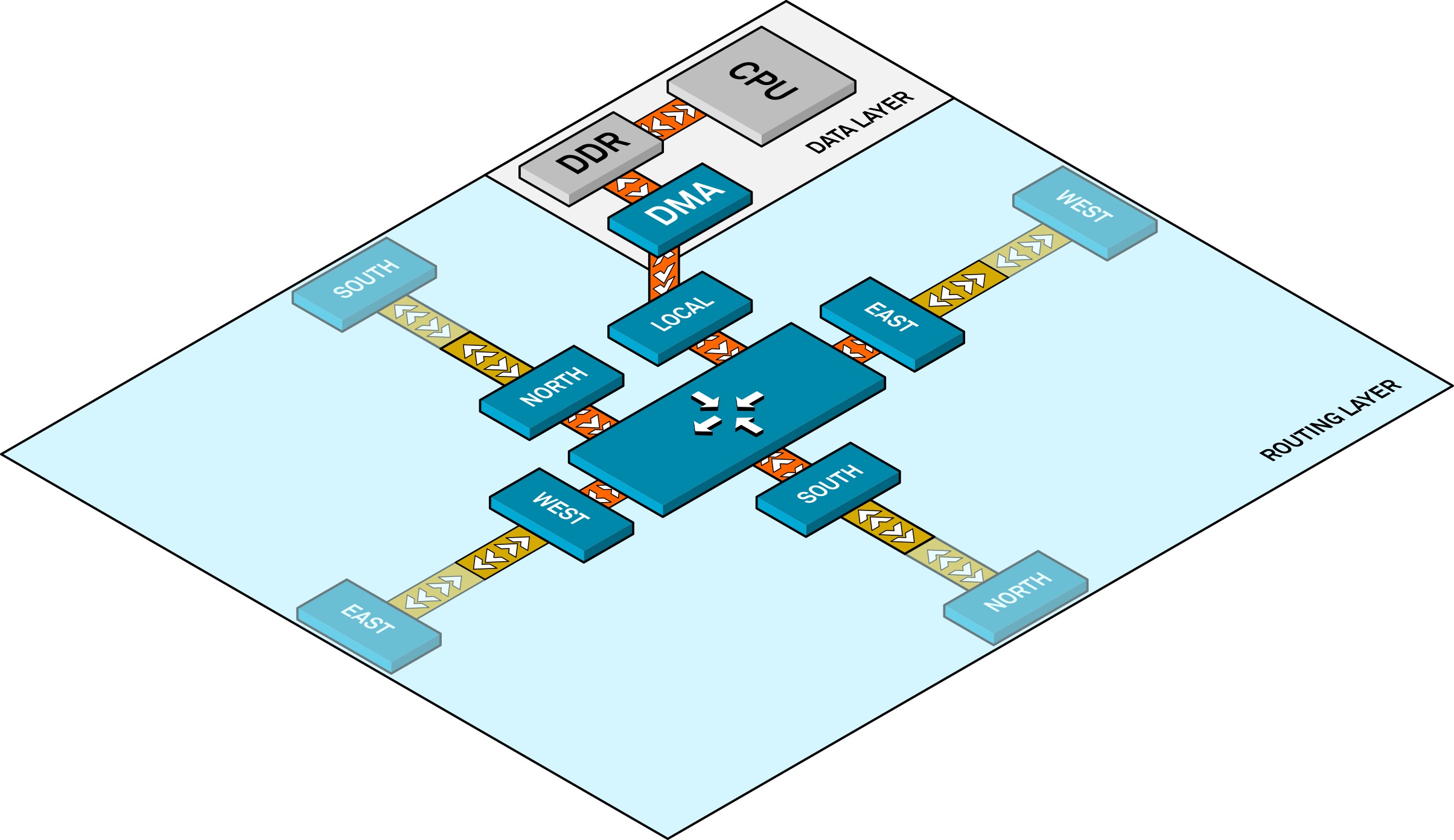}
  \caption{SCALP node internal architecture for a 2D array.}
  \label{fig:scalp_router}
\end{figure}%

\begin{itemize}
\item 
The routing layer includes the equivalent of routing + link + physical layers in the OSI model, it guarantees link integrity between two neighbor nodes, and can redirect packets to neighbors in order to permit remote node communications without using intermediate computation nodes. A set of crossbar switches included in the FPGA supports the packet transmission.   
\item The computation layer is mainly composed of the processor running the application, its memory and a Direct Memory Access (DMA) channel. The processor is in charge of handling application data in its local DDR memory and configuring DMA transfers for sending data to the routing layer. The interface between the router and the DMA is the same as the one between neighbor routers, built as an AXI stream interface. This uniform standard interface ensures the scalability of the system, in order to allow it to evolve to eventually add heterogeneous computation nodes.
\end{itemize}

SCALP is thus an excellent candidate for implementing ReSOM because its inherent cellular architecture permits a scalable deployment, given the distributed computing paradigm and the lack of hardware inter-dependency between nodes. Each of the SOM modalities can evolve in an independent SCALP board that adapts resources to its own computation needs. The communication is made transparent whatever the location of the connected maps, either local neighbor or remote. The correlation between SOMs can thus be performed in a distributed and potentially asynchronous manner.


\section{Results}
\label{sec:Results}

\subsection{Simulation results}

\subsubsection{ReSOM framework and dataset description.}

To test the model, a Python-based framework was implemented using the PyTorch library \cite{NEURIPS2019_9015} to speed up matrix calculations. Some simple datasets are used for the tests, such as the MNIST \cite{lecun_mnisthandwrittendigit_2010} and Spoken-MNIST (SMNIST) datasets. The SMINST dataset is a subsample of Speech Commands \cite{warden2018speech} reduced to pronounced numbers from 0 to 9. The numbers were transformed to the Mel Frequence Cepstral Coefficients. The final version of the joint dataset in this configuration for two modalities was presented in a previous work \cite{Khacef2020_MultimodalData}.

The Fashion MNIST dataset \cite{xiao2017fashionmnist}, representing images of different clothes, and the hand signs dataset \cite{mavi2021new}, representing numbers shown by hands, were used as the third and fourth modalities. 
Even if F-MNIST does not represent the same type of information as other considered datasets, it is one of the simplest datasets in the literature and it uses the same format as MNIST (10 classes and 28x28 images). It is assumed that each of the objects present in this dataset is arbitrarily associated to a specific class number between 0 and 9.
More information about the dataset size can be found in Table \ref{table:datasets sizes}.

All the datasets have been subsampled or oversampled (if needed) to have the same number of training/test vectors (60 000/10 000), but the data vector size is left unchanged. All the vectors have sizes comprised between 100 and 2000 depending on the dataset used. We applied our labeling algorithm on a part of 5000 training data vectors (about 8\% of data were labeled).

\subsubsection{Preliminary tests of ReSOM with 2 modalities}
Due to the use of a new framework that accelerates matrix calculations, it was decided to return to the problem of 2 modalities, previously resolved with ReSOM. We check if the model improves with a wider search for the hyperparameters, keeping the old sizes of the SOMs (\(10\times10\) and \(16\times16\) for MNIST and SMNIST, respectively). This time instead of a hard grid for such parameters as \(\sigma\) and \(\epsilon\), an advanced algorithm "Optuna" \cite{akiba2019optuna} was used to optimize their values.

As can be seen in Table \ref{table:2xmodal_acc}, the accuracy results of the previous similar work (95.1\%) have been achieved and even surpassed (with 96.6\%). Notice that our model has used the datasets in a raw form, without any feature extractor, which might radically enhance the quality of the vectors representing the data. The addition of feature extractors as input of the neural maps increases the algorithm's accuracy, as we showed in \cite{Khacef2020_FewShotGPU, Khacef2020_FeatureSOM}.

\subsubsection{Tests of ReSOM scalability to more modalities}
The following tests presented in Table \ref{table:4xmodal_acc} show the model's scalability. We conducted tests of ReSOM with up to four modalities by adding the Fashion MNIST database (FMNIST) and a hand gesture database (denoted "Gests" in the following). It can be seen that the addition of new modalities can significantly increase the accuracy (up to the 99\% in our tests). The best performance was achieved after optimizing the hyperparameters, details of which are available in a table in the article's additional materials. 

\begin{figure}[ht]
    \centering
    \includegraphics[width=0.5\linewidth]{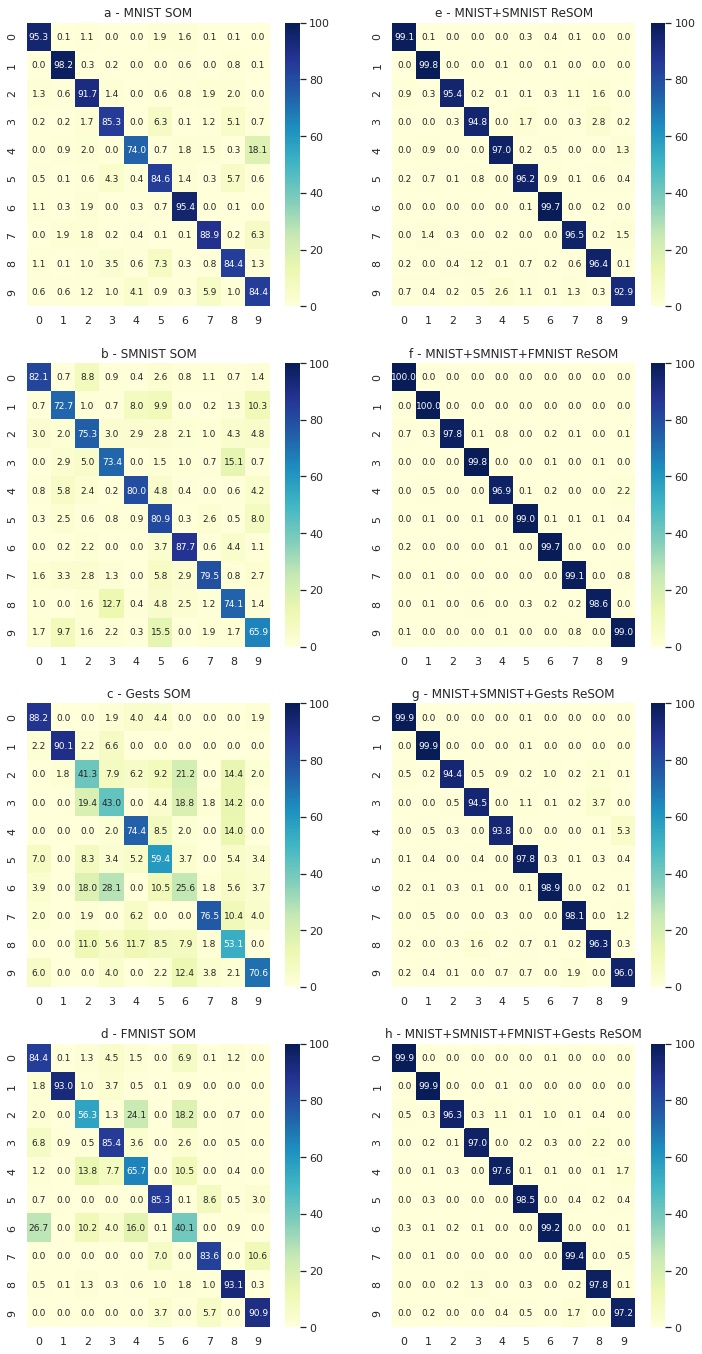}
    \caption{The first column \textbf{(a-d)} is confusion matrices for SOM predictions for 4 datasets: MNIST, SMNIST, Gests and FMNIST. The second column \textbf{(e-h)} is ReSOM results for different combinations of datasets: "MNIST+SMNIST", "MNIST+SMNIST+FMNIST", "MNIST+SMNIST+Gests" and "MNIST+SMNIST+FMNIST+Gests".}
    \label{fig:Conf_martix}
\end{figure}

A deeper analysis may be done by analyzing the confusion matrices of each neural map (Fig. \ref{fig:Conf_martix}). The use of the more modalities means each one helps to compensate for some of the mistakes of others. As an example let's look at the prediction accuracy for the number "9". The prediction accuracy of "MNIST + SMNIST" (Fig. 6e) is relatively low (92\%). But it reaches close to 99\% (Fig.6f) using the extra capacities of FMNIST (Fig. 6d), a modality which is quite good at predicting the number "9". At the same time, adding the "Gests" modality (Fig.6c) was not as useful (Fig. 6g - 96\%), possibly because of its lower accuracy in predicting "9" compared to the FMNIST modality. 

Also we note that having greater number of modalities may even decrease the result (Fig. 6h), either because of the addition of incompatible information (for lack of an adequate database) or due to the complexity of searching for the optimal hyperparameters of the model.

\subsubsection{Connection importance and pruning}
\begin{figure*}[ht]
    \centering
    \includegraphics[width=\linewidth]{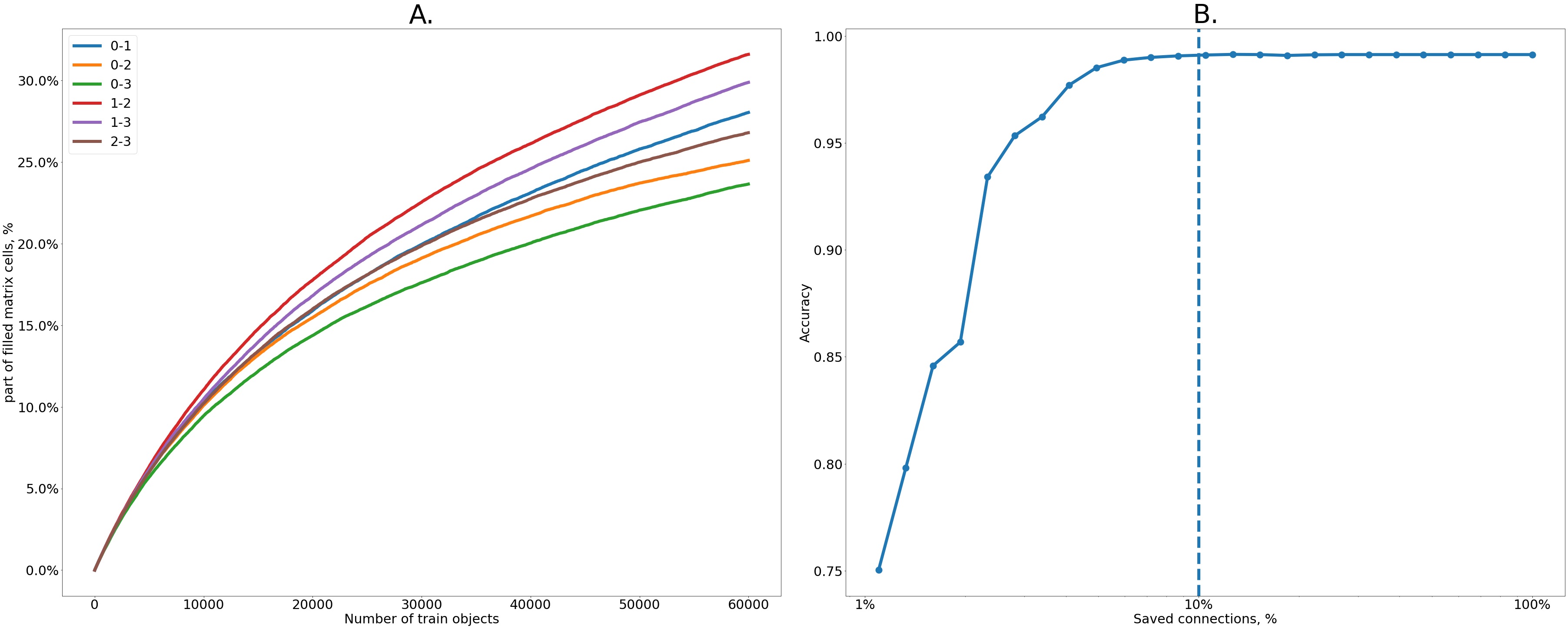}
    \caption{On the left \textbf{(A)}: evolution of the fraction of nonzero lateral connections between pairs of SOMs during learning (all SOMs are fixed to a size of 16x16): 0 - MNIST, 1 - SMNIST, 2 - Gests database, 3 - FMNIST. On the right \textbf{(B)}: How the accuracy changes with the rate of kept connections (100\% - all the connections are kept). The abscissa is in log scale.}
    \label{fig:Resom_connections}
\end{figure*}

Reducing the number of values in the matrix can be extremely useful both for speeding up calculations and for reducing energy consumption in future implementations of the system. So next, we analyze the level of sparseness of the  connections obtained and its influence on the resulting accuracy. The first thing to look at is the number of unused connections (i.e., those with zero weight). The fraction of such connections is quite large, as can be observed in Fig. \ref{fig:Resom_connections}A, which shows the evolution of the number of nonzero connections during the training process.

We can observe some differences in the density of connections depending on the SOMs to be linked together. For example, the difference in the number of connections about of \(1.3\) times between the most linked pair of SMINST and Gests database ("1-2"), and the less linked pair of MNIST and FMNIST ("0-3"). The difference in connectivity may, for example, be determined by the complexity of the represented modalities, which, in turn, may influence the frequency of activation of various neurons. 

Also, we can observe that the training has still not reached a plateau - therefore, the limit of the created connections can be much higher than the 30\% shown in Figure \ref{fig:Resom_connections}A. Thus we can suppose that many more neurons may become connected over time. This level of sparsity makes storing the matrix in a compressed format not profitable and quite unjustified. But we still wish to avoid the unnecessary memory usage, so we explore how strongly we can increase the matrix sparsity without a decrease in accuracy.

In fact, not all of the combinations have the same weight, so the influence of some might be much less than others. This leads us to the idea of cutting the weakest connections, that is, those attached to the smallest weight values. To search for the optimal threshold we track the influence of the number of smallest weights on the model's accuracy, equating to zero a certain percentage of the weakest lateral connections. This information is plotted in Figure \ref{fig:Resom_connections}B. As long as more than 10\% of the nonzero connections are retained, there is no significant effect on the accuracy. Thus about 3\% of matrix cells (10\% of the average 30\% cells filled among all matrices) are enough to achieve the maximal model accuracy.

\subsection{Results of the deployment of the model on the hardware platform}

\subsubsection{General description of the experiment scheme}
\begin{figure}[ht]
    \centering
    \includegraphics[width=0.7\linewidth]{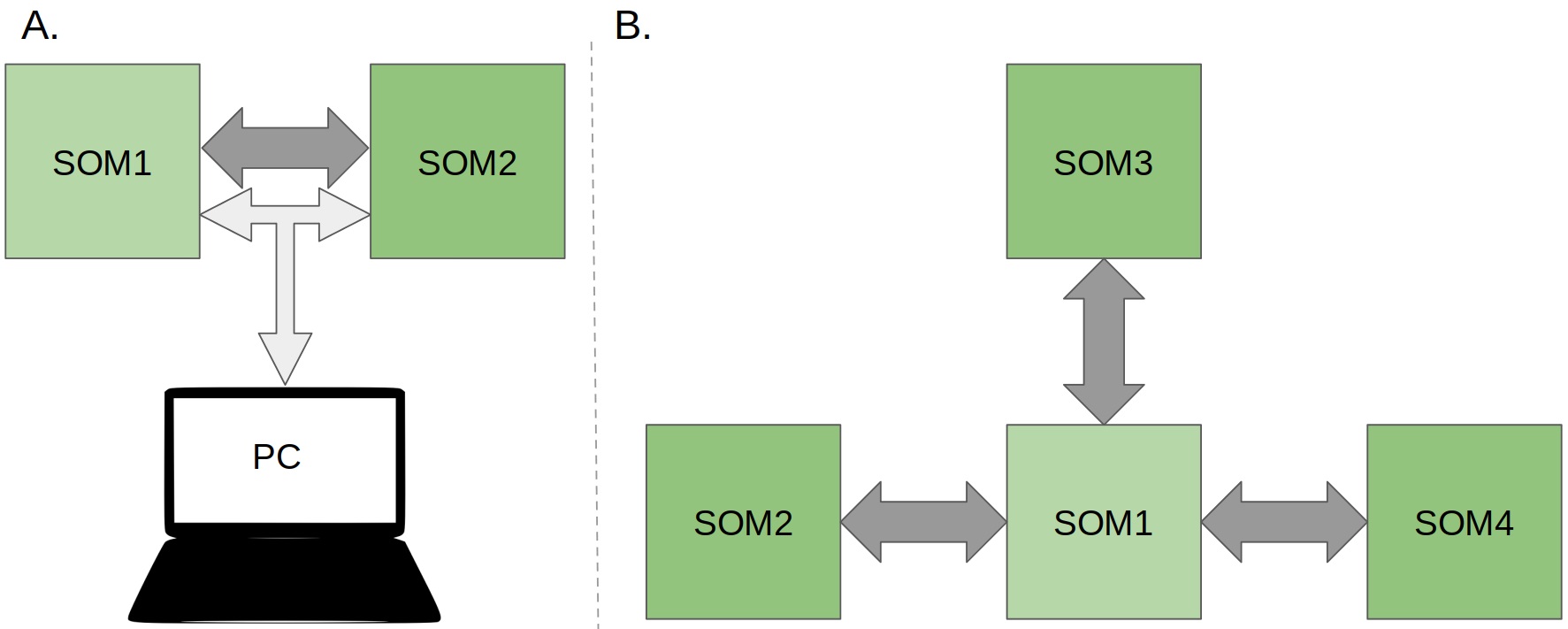}
    \caption{On the left (A): Scheme of the deployment of the ReSOM model onto the SCALP electronic boards for 2 modalities. (B): ReSOM + SCALP processing 4 modalities. Dark grey and light grey arrows represent respectively SCALP high speed serial connections and an ethernet LAN network. The PC is not showed in the second image for the sake of simplicity.}
    \label{fig:SOMs_connections}
\end{figure}

To demonstrate the feasibility of the algorithm on real hardware, a simple scheme of interaction of two SOMs was implemented on the SCALP boards. The scheme involves 2 SCALPs passing data directly via the HSSL protocol. They are responsible for processing visual and auditory information and one of them also for the ReSOM prediction inference. The boards are connected to a PC in a local network via an Ethernet connection. The scheme can be seen in Figure \ref{fig:SOMs_connections}A.

In the proposed scheme, the PC plays the role of program execution controller. It starts the different steps of the algorithm and controls the synchronization of data transfer between the devices. It also sends data vectors for testing and collects the final predictions, playing both the role of sensors (such as a camera) in a real system and the role of a monitoring system for the counted predictions. 

Some parts of the algorithm (such as the calculation of SOM activations and the modification of the weights while training) are parallelized and so executed on different boards and computed independently of one another. These processes are controlled in different threads of the controller program written in Python and launched on the PC. The SOM calculations are executed on the boards as a Python server program. The nodes expose SOM and ReSOM computing functions as services that can be called through Python's RPyC library. All requests are sent over the internal LAN network configured for standard communication between Bionic Beaver Linux on the PC and PetaLinux on the SCALP board.

For communication between the boards, direct connections with the HSSL protocol are used. Its schema allows the address of a node in the network to be set in three-dimensional coordinates and data to be sent to a desired address. In the current SCALP version, data are sent on a FIFO receiver of a board, where data packets of a size of 64 words of 8 bytes are sequentially recorded. The packet sending function is a core application service and may be called via C/C++ code (or by a Ctypes call in Python). To send a data object of any type and size in Python, a program wrapper was written that serializes and splits files into the packets, and after that sequentially sends them to a given address. The receiver board reassembles them to the initial format. 

\subsubsection{Deployment scenarios}
Several deployment scenarios were tested to verify the global system performance. The first model to be tested was the inference model, to demonstrate the ability of a two-board system to produce values similar to those simulated earlier, using the already trained weights of both SOMs and lateral connections. The schema followed is shown in Table \ref{table:scalp_resom_communication_pipeline}.

Further, in a similar way, the model was tested with the addition of the training of ReSOM lateral connections. The pipeline of information exchange is the same as the one presented earlier, except that it also contains an additional training step (see Table \ref{table:scalp_resom_communication_pipeline}). 

\subsubsection{Verifying the functioning of the hardware framework}
This test consists of verifying the model's functioning on the hardware platform by comparison with the results of the previously tested PC simulation. The training of the entire ReSOM system is assumed to have been done offline on a PC. The weights are then fixed and the multimodal data are sent over the network to each neural map. So we compare two inference implementations to see if they work the same way on the same data. 

Achieving the maximum model performance was not the focus of the research in this paper; the main goal of the tests was to demonstrate whether such a model could be implemented. Because of some nonoptimal features of the presented framework (which will be discussed later), we did not reach the maximum possible computational speed. Therefore we decided to limit ourselves to a small test dataset (300 samples), sufficient to demonstrate the model's functionality.

\begin{figure}[ht]
    \centering
    \includegraphics[width=0.5\linewidth]{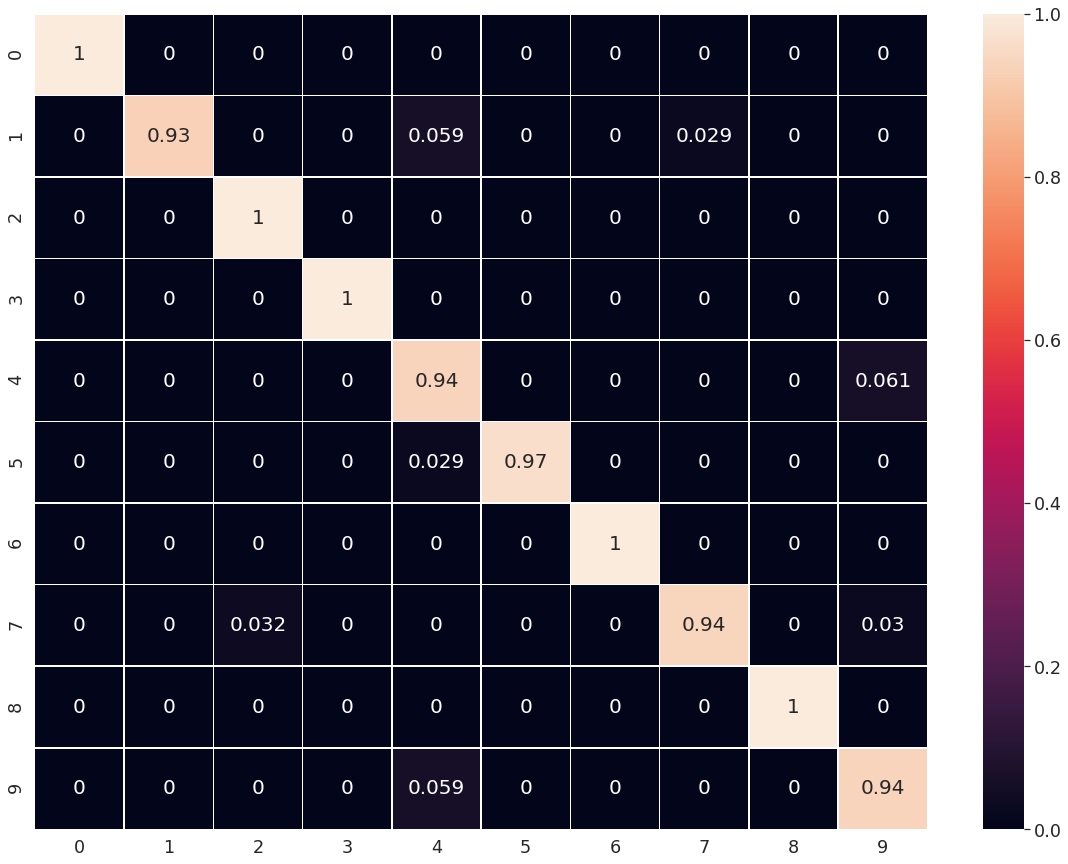}
    \caption{Confusion matrix of ReSOM on tests conducted on SCALP boards. Accuracy = 0.966, test dataset size = 300 pairs of MNIST/SMNIST objects.}
    \label{fig:ReSOM_scalp_conf_matr}
\end{figure}

The accuracy of this on-board prediction test was equal to 96.6\% and the resulting confusion matrix is presented in Figure \ref{fig:ReSOM_scalp_conf_matr}. These results are exactly the same as the ones achieved by the simulation code, both in term of accuracy and the confusion matrix, thus confirming the validity of the deployed model.

\subsubsection{Analysis of system execution performance}
In the previous section, we conducted tests with a fairly small number of samples. This was due to the the hardware architecture of the FPGA and non-optimized usage of the Python language. So the on-board execution for this first prototype does not yet reach the computation speed expected. But the use of multiple boards gives us the ability to perform parallel computations over the boards, which already gives a significant gain in system performance. The following section provides an analysis and numerical evaluation of the possible gain when scaling the system to more than two modalities. Another level of parallelism consists in distributing the SOM map inside the FPGA circuit thanks to the Iterative Grid algorithm. This optimization will be considered in prospective work. 
As presented in Table \ref{table:times_of_scalp_execution}, the calculation time of the SMNIST (neural map) activation matrix is an order of magnitude slower than its transfer time. Thus, computing in parallel and then transferring will result in a gain in system execution time over computing the algorithm on a single device. Adding more devices will enable further parallelization gains and therefore a greater increase in performance relative to executing on a single board. To demonstrate this, consider a model system composed of 4 SOMs implemented on 4 boards (Fig.\ref{fig:SOMs_connections}B).

In this case, the parallel solution takes 0.208 s compared with 0.64 s for one board (as shown in Table \ref{table:times_of_N_scalps_execution}); so the system gains a factor of 3 in the time needed to calculate and prepare the neuron activation matrices. This speedup is proportional to the number of boards in use. But we should note that the acceleration rate might decrease with the system's growth and may reach a state where it is impossible to directly connect all the boards to the one responsible for the ReSOM calculation.

\subsection{Conclusion}
\subsubsection{Brain-inspired model}
The cerebral cortex is capable of using self-organization for learning in the environment without the presence of embedded annotations. Individual cortical areas are responsible for processing different signals (in other words, modalities). Interaction between cortical zones allows the brain to build a complete picture of the world, reflecting a complex picture of interactions between different modalities using local computations.

This work combines simplified biological models to propose an architecture capable of similar behavior. ReSOM enables self-organization of the system and aids lateral connections between the "cortices" to exchange information, correcting weaknesses and information leaks among the different modalities. Our model is based on an interpretation of the interaction between real cortical zones. This fits perfectly with the idea of building a scalable system. Such a system was naturally implemented on microprocessor boards, allowing independent operation of several computational elements and improving the system's performance.

\subsubsection{Implementation}
The architecture allows the number of modalities to be scaled, and the computational efficiency will grow non-trivially by adjusting the number of SOMs. The model reaches an accuracy of 99\% using only 3\% of all possible ReSOM connections. This addresses a specific ML problem - the post-labeled unsupervised learning problem. Its advantage is the ability to determine the predicting system almost without any labels, partially solving the problem of expensive annotation.

The proposed model is implemented in hardware on SCALP boards. The possibility of scaling up the system to several modalities is also demonstrated. The use of high-speed serial connections allows information to be transferred directly between the boards, which gives an acceleration in system performance.


\section{Discussions}
\label{sec:Discussion}

\subsection{Limitations of the current prototype}

\subsubsection{Growth of the number of lateral connections}
The number of lateral connections, and the underlying Hebb computations, is a very important parameter that affects the memory consumption and execution time. To estimate the number of lateral connections for the ReSOM model, we can derive the following formula (\ref{eqn:num_of_lat_conn_resom}), which has a quadratic growth rate.

\begin{equation}
C_{ReSOM} = C_n^2 = \frac{n(n-1)}{2};
\label{eqn:num_of_lat_conn_resom}
\end{equation}

For example, in the case of 7 boards the ReSOM architecture will have 21 lateral connections, that is, 3 times more than the number of boards. So, for a small number of modalities, this growth is not so important, but as the number of modalities increases, it may affect the complexity of the signal transmission or the energy consumption. 

An alternative method of multimodal aggregation is the convergence-divergence zone (CDZ)-based SOM combining all modalities in a single CDZ map. This method is not discussed in detail in this paper, though we mentioned it in the literature review. To our knowledge, the ReSOM model currently offers higher accuracy than the CDZ. But, with a linear growth rate in the number of lateral connections, the CDZ-based SOM is a worthy candidate for developing a more scalable system, achieving a more impressive speed gain for a greater number of modalities.

\subsubsection{Python overhead}
To speed up the prototype production and to simplify further integration with ready-made AI libraries, a significant part of the framework's code is written in Python. Unfortunately, non-optimized Python code entails a significant computation inefficiency and increase in execution time. This is largely due to the need to serialize, split into packets and then assemble the counted activations.

Such a decrease in the speed is not critical when the activation computation time significantly exceeds the transmission time. But in future implementations, with an increase in the calculation speed, this process will need to be optimized. This can be achieved either by rewriting the serialization-deserialization Python code in a low-level language as C/C++, or by reducing the limit on the size of forwarded packets. To solve this problem, we plan to implement a DMA interface between directly connected boards in a future SCALP system update.

\subsection{Possible development directions}
\subsubsection{Hardware development}
A distinctive feature of SCALP is an integrated FPGA chip, which enables a grid of independent computing elements to be built. This paper ignores this feature of the boards, but some solutions have previously been proposed \cite{girau2019cellular, Khacef2020_PhDthesis}. They develop the idea to use the FPGA for executing a SOM and accelerating the algorithm for calculating its BMU. Due to the direct exchange of information between neighboring SOM neurons, we can reduce the algorithm's complexity from \(O(n)\) to  \(O(\sqrt{n})\), where \(n\) is the number of SOM neurons. In future, it is planned to correct this drawback and integrate one of the previously proposed Iterative Grid \cite{rodriguez2018ig_som} or Cellular SOM \cite{girau2019cellular} solutions in order to significantly speed up the calculations.

At this stage, in order to keep up with the speeding up of the SOM computations, the DMA implementation is much more important and useful, therefore this modification is already planned for the next SCALP release and will be integrated into our architecture.

\subsubsection{Next technical algorithm deployment}
The obvious next step consists in testing the architecture's operation in the presence of 4-5 boards for processing different data modalities and comparing the performance of the ReSOM and CDZ methods for aggregating multimodal data in terms of accuracy, latency and power consumption. Growing the system up to more than a dozen boards makes sense to increase its performance, similar to the multi-thread solution proposed by \cite{jayaratne_2021}. 

A further increase in the number of data modalities will be slowed down due to the difficulty of finding such a large number of information channels of a different nature. A possible way to solve this problem is to allow some of the boards to work with data of the same type, creating independent SOMs with data of the same nature. The development of such architectures will require both adaptation of the algorithm to work with several data aggregation nodes and new technical solutions for the implementation of the structure. This innovation could significantly boost the model's performance, so this extension is also planned for future implementations.

\subsubsection{Online learning, acting and spiking neural networks}
The ability to process signals and act in real time is an important feature of living systems that we have not discussed in this work. In future work, we will address aspects of our architecture that will allow it to develop into a full-fledged online acting agent.
Some previously cited works \cite{lallee_multi-modal_2013, 10.3389/frobt.2016.00022, lefort2010self, morse2010epigenetic} show a significant potential for the development of self-organizing maps as a method for robots to navigate the surrounding space. The maps are able to model the spatial movement of agent and objects in space, so the multimodal connection can also help to capture the nature of spatial phenomena.

This model has no restrictions for processing information online and can also be used for real-time processing of incoming signals. This could be done by integrating with spiking neural networks (SNNs). SNN have a distributed network structure \cite{zora75356, 938430, ghosh-dastidar_third_2009}, but they do not forget the signal nature of received data and process them as spikes, in a sequential mode. Algorithms using SNNs have already allowed us to solve quite important and varied problems, such as unsupervised learning \cite{dong_unsupervised_2018,991428}, auto-encoding \cite{kamata2021fully} and even supervised AI problems \cite{Kheradpisheh_2020}. Also, their good performance in terms of prediction accuracy and energy consumption \cite{8879613,kim2019spikingyolo} promise their great potential for further development. The concept of SSN can be also combined with the SOM, as shown by other researchers \cite{Hazan_IJCNN_2018}. Thus we see great potential to develop our architecture by integrating it with SNNs for signal-type data processing.

The ReSOM multimodal association learning methods explored in this work were performed sequentially in two phases: first, we trained the SOMs for unimodal classifications, and second we created and reinforced bidirectional connections between pairs of maps based on their activities on the same training dataset.
We refer to this learning approach as \textit{asymmetric}.
This is particularly interesting in the context of offline learning when working on pre-established datasets.
First, in a purely practical way, it gave a lot of flexibility since we could train the unimodal SOMs on their respective available data separately, then train the multimodal association based on a smaller synchronized multimodal dataset.
Synchronized here means that the multimodal samples that belong to the same class are presented at the same time.
Second, from a developmental point of view, it has been shown that auditory learning begins before birth while visual learning only starts after birth \cite{althaus2013modeling}. 
Moreover, the ability to build associations between words and objects in infants appears to develop at about 14 months of age \cite{werker1998word_object_association}.
The opportunity to process visual and auditory information sequentially may offer computational advantages in infant learning, as it could be a facilitating factor in the extraction of the complex structures needed for categorization \cite{althaus2015timing_matters}.
These observations support the actual learning approach of ReSOM, where multimodal associations begin to develop after unimodal representations are learned sequentially.

Nevertheless, in the context of online learning in a dynamic and changing environment, another approach would be to perform both Kohonen-like and Hebbian-like learning at the same time, continuously. 
For example, this approach is followed using STDP learning in \cite{rathi2018multimodal_stdp}.
For this purpose, the KSOM would be replaced by a DSOM. The reason is that the KSOM has a decaying learning rate and neighborhood width, so that the learning stabilizes after a certain number of iterations. Therefore, the learning is stable but not dynamic, and can be considered as an offline unsupervised learning algorithm. In contrast, the DSOM is a variation of the KSOM algorithm where the time dependency of the learning rate and neighborhood function has been replaced by a dependence on the distance between the BMU and the input stimulus. While the DSOM is less accurate than the KSOM \cite{Khacef2019_SOMs}, it is more suitable for online learning.
In addition, we would need a dynamic learning rate so that the multimodal association becomes stronger when the sample is well learned by the SOM, i.e. when the distance between the BMU and the sample is small. One way to do that is to use Gaussian kernel-based distances, so that the multimodal binding becomes more relevant after the convergence of the SOMs, without any manual tuning of the SOM hyperparameters.

\subsubsection{Application to other problems}
This work proposes one possible method for using lateral connections to transfer activations between SOMs. The model's evaluation occurs using a non-negligible number of annotations (at least 1 \% of the training data). This is somewhat in conflict with the proposed self-organizing model, which is capable of completely unsupervised reasoning. However, the information stored in lateral connections may be rich enough to define the stable clustering on its own, with almost no use of the labels. Therefore, it seems possible to develop a completely unsupervised, or much less labeled algorithm, using graph cutting or distance clustering methods.

Such a system might be capable of learning on real data (such as video and audio signals captured simultaneously) by creating the clustering for all signal modalities using only the dependencies between them. Thus, simply by observing the objects around it, a robot could be capable of dividing the world into separate classes or categories. The model will thus be able to "understand" the world, by learning its distinguishable concepts. The development of such models is planned for future research studies.

\section*{Conflict of Interest Statement}

The authors declare that the research was conducted in the absence of any commercial or financial relationships that could be construed as a potential conflict of interest.


\section*{Author Contributions}

A.M. made the model's simulation on a PC and implemented tests of the model on the SCALP electronic boards. He also wrote the majority of the text.

L.R. implemented the library of information exchange between electronic boards in Python, actively participated in the discussions on the direction of the work.  He wrote some parts of the text and contributed significantly to the editing.

B.M. supervised the whole process of the work preparation, took an active part in the discussions about the direction of the work. He participated extensively during article's writing.

L.K. did preliminary tests of the model and took part in the planning of SCALP experiments. He also contributed to the article's writing and editing.

J.S. actively participated in the design of the SCALP architecture, implemented the installation of the board's system and the programming of the messaging protocol between the boards.

Q.B. actively participated in the development of the SCALP architecture, configured the communication system between the boards.

A.U. supervised the implementation of the hardware part of the work. He took part in the discussions about the direction of the work, wrote some of the text and participated in editing the article.


\section*{Funding}
This work has been supported by the French government, through the 3IA Côte d’Azur Investments in the Future project managed by the National Research Agency (ANR) with the reference number ANR-19-P3IA-0002.

It has been also supported by the Swiss National Science Fund (SNSF) through the international SOMA project with the reference ANR-17-CE24-0036.



\section*{Acknowledgments}

Special thanks to Madame Catherine Buchanan, Scientific Editor at Université Côte d'Azur, for her meticulous work in correcting the text and improving its writing style.



\section*{Data Availability Statement}

The datasets analyzed in this study can be found in : MNIST and Speech Commands - \cite{Khacef2020_MultimodalData}, FMNIST  - \cite{xiao2017fashionmnist}, and hand signs dataset - \cite{mavi2021new}.



\section*{Tables}

\begin{table}[!ht]

 \label{tab:datasets_sizes}
 \caption{Characteristics of the datasets used in the study.}
 
\begin{tabular}{|p{0.4\textwidth}|p{0.15\textwidth}|p{0.15\textwidth}|p{0.15\textwidth}|}
 \hline
 \multicolumn{4}{|c|}{Datasets table} \\
 \hline
 DataSet & Vector size& Unique training objects & Unique test objects\\
 \hline
MNIST \cite{lecun_mnisthandwrittendigit_2010}  & 28x28 & 60000 & 10000\\
Spoken MNIST \cite{warden2018speech} & 39x13 & 34801 & 4107\\
Fashion MNIST \cite{xiao2017fashionmnist}  & 28x28 & 60000 & 10000\\
Hand signs \cite{mavi2021new} & 28x28 (reshaped from 64x64) & 1531 & 515\\
 \hline
\end{tabular}

 \label{table:datasets sizes}
\end{table}

\begin{table}[!ht]
 \caption{Accuracy comparison of different methods processing 2 modalities.}
\begin{tabular}{|p{0.8\textwidth}||p{0.15\textwidth}|}
  \hline
 Test name& Accuracy(\%)\\
 \hline
 \multicolumn{2}{|c|}{Unimodal} \\
 \hline
MNIST (STDP SOM \cite{rathi_stdp_2021} ) & 93.2 \\
TI46 (stream audio) (STDP SOM \cite{rathi_stdp_2021} ) & 96.0 \\
MNIST (Convolutional clustering unsupervised, 3000 objects for labeling \cite{dundar_convolutional_2016}) & 98.6 \\
MNIST (Deep Semi-supervised SOM, 10\% of labels \cite{braga2020deep}) & 97.4 \\
MNIST (Unsupervised STDP \cite{diehl_unsupervised_2015}) & 95.0 \\
MNIST (SOM - 10x10, this work)  & 88.4 \\
Spoken-MNIST or SMNIST (SOM - 16x16, this work) &   77.0 \\
\hline
\hline
 \multicolumn{2}{|c|}{2 modalities} \\
\hline
MNIST+TI46 (audio) (STDP Multimodal SOM \cite{rathi_stdp_2021} ) & 98.0 \\
MNIST+SMNIST (Khacef et al. ReSOM \cite{Khacef2020_ReSOM} ) & 95.1 \\
 \hline
MNIST+SMNIST (ReSOM - this work) & 96.1 \\
MNIST+SMNIST (ReSOM + optuna - this work) & 96.6 \\
 \hline
\end{tabular}
 \label{table:2xmodal_acc}
\end{table}

\begin{table}[!ht]
 \caption{Comparison of accuracy up to 4x modalities with the ReSOM architecture.}
\begin{tabular}{|p{0.4\textwidth}||p{0.15\textwidth}|}
 \hline
 \multicolumn{2}{|c|}{Conducted tests} \\
 \hline
 Test name& Accuracy(\%)\\
 \hline
 MNIST SOM (10x10)  & 88.4 \\
 Spoken-MNIST (16x16) &   77.0 \\ 
 Fashion MNIST SOM (10x10)  & 78.47 \\
 Gests SOM (16x16) &   55.63 \\
 MNIST+SMNIST & 96.6 \\
\textbf{MNIST+SMNIST+FMNIST} & \textbf{99.0} \\
 MNIST+SMNIST+Gests & 97.2 \\
 MNIST+SMNIST+FMNIST+Gests &  98.36\\
 \hline
\end{tabular}
 \label{table:4xmodal_acc}
\end{table}

\begin{table}[!ht]
 \caption{Board interaction pipeline for the ReSOM inference.}
\begin{tabular}{|p{0.3\textwidth}||p{0.3\textwidth}||p{0.3\textwidth}|}
 \hline
 PC & Board 1 & Board 2 \\
 \hline
 \hline
 Ask 2 boards to init themselves & -  & - \\
 \hline
 - & Load MNIST weights, labels and ReSOM weights & Load SMNIST weights and labels \\ 
 \hline
 Confirm 2 boards init  & -  & - \\
 \hline
 \multicolumn{3}{|c|}{Start of loop} \\ 
 \hline
 \hline
 Send vector 1 to SOM1 and vector 2 to SOM2 &   -  & - \\
 \hline
 - & Count activations of MNIST map  & Count activations of SMNIST map \\
 \hline
 Confirm end of counting  & -  & - \\
 \hline
 Ask board 2 to send activations to board 1 & -  & - \\
 \hline
 - & receive SMINST activations  & send SMINST activations \\
 \hline
 Confirm data received  & -  & - \\
 \hline
 Ask board 1 to make a prediction using ReSOM model& -  & - \\
 \hline
 - & Count ReSOM predictions using 2 activation maps  & - \\
 \hline
 Ask for and get the label prediction & - & - \\
 \hline
 \hline
 \multicolumn{3}{|c|}{End of loop} \\ 
 \hline
 Compute final accuracy and confusion matrix & - & - \\
 \hline
\end{tabular}

 \label{table:scalp_resom_communication_pipeline}
\end{table}

\begin{table}[!ht]
 \caption{Average execution times of inference steps.}
\begin{tabular}{|p{0.22\textwidth}|p{0.5\textwidth}|p{0.18\textwidth}|}
 \hline
 Step \# & Algorithm step & time, s.\\
 \hline
 1 & Count MNIST activations (SOM1, 10x10 size) & 0.0795 \(\pm\) 0.005 \\
  \hline
 2 & Count SMNIST activations (SOM2, 16x16 size) & 0.1550 \(\pm\) 0.005 \\
  \hline
 3 & Send activations from one board to another (matrix of 16x16 numbers) & 0.016 \(\pm\) 0.001 \\
  \hline
 4 & Make ReSOM prediction using 2 activations & 0.011 \(\pm\) 0.001 \\
 \hline
 \hline
 max(\#1,\#2) + \#3 + \#4 & \textbf{SCALP ReSOM on 2 boards} & \textbf{0.182 \(\pm\) 0.01} \\
 \hline
 \hline
 \#1 + \#2 + \#4 & SCALP ReSOM on 1 board (calculated from the measured times) & 0.246 \(\pm\) 0.01 \\
 \hline
\end{tabular}
 \label{table:times_of_scalp_execution}
\end{table}

\begin{table}[!ht]
 \caption{Gain in execution time for scaled system.}
\begin{tabular}{|p{0.2\textwidth}|p{0.5\textwidth}|p{0.2\textwidth}|}
 \hline
 Act. \# & Alg. step & time, s.\\
 \hline
 1 & Count activations (approximate, same for all boards with same SOM of 16x16 cells) & 0.16 \\
  \hline
 2 & Send activations from one board to another (hypothetical) & 0.016 \\
  \hline
 \hline
 \multicolumn{3}{|c|}{\textit{Data preparation for N modalities ReSOM}} \\
 \hline
 \hline
 (\#1)+(n-1)*(\#2) & \textbf{SCALP ReSOM on N boards} & 0.16+(n-1)*0.016 \\
 \hline
 \hline
 n*(\#1) & SCALP ReSOM on 1 board &  n*0.1 \\
 \hline
 \hline
 \multicolumn{3}{|c|}{\textit{Data preparation for 4 modalities ReSOM}} \\
 \hline
 \hline
 (\#1)+3*(\#2) & \textbf{SCALP ReSOM on 4 boards} & 0.208 \\
 \hline
 4*(\#1) & SCALP ReSOM on 1 board &  0.64 \\
 \hline
 
\end{tabular}

\label{table:times_of_N_scalps_execution}
\end{table}


\bibliographystyle{IEEEtran}
\bibliography{citations}

\end{document}